\documentclass[11pt,letterpaper]{article}
\usepackage{naaclhlt2015}
\usepackage{times}
\usepackage{latexsym}
\usepackage{comment}
\usepackage[T5,T1]{fontenc}
\usepackage{comment}
\usepackage{algorithmic, amssymb, amsmath}   
\usepackage{algorithm}
\usepackage{multirow}
\usepackage{devanagari}
\usepackage{amsmath}
\usepackage{multirow}
\usepackage{color}
\usepackage{wrapfig}
\usepackage{array}
\usepackage{graphicx}
\title{Phrase translation using a bilingual dictionary and n-gram data: \\A case study from Vietnamese to English}

\author{Khang Nhut Lam, Feras Al Tarouti and Jugal Kalita \\
Computer Science Department \\
University of Colorado, Colorado Springs, USA \\
{\tt \{klam2,faltarou,jkalita\}@uccs.edu}\\}  

\date{}

\begin{document}
\maketitle
\begin{abstract}
Past approaches to translate a phrase in a language $L_1$ to a language $L_2$ using a dictionary-based approach require grammar rules to restructure initial translations. This paper introduces a novel method without using any grammar rules to translate a given phrase in $L_1$, which does not exist in the dictionary, to $L_2$. We require at least one $L_1$--$L_2$ bilingual dictionary and n-gram data in $L_2$. The average manual evaluation score of our translations is 4.29/5.00, which implies very high quality. 
\end{abstract}

\section{Introduction}

This paper tackles the problems of phrase translation from a source language $L_1$ to a target language $L_2$. The common approach translates words in the given phrase to $L_2$ using an $L_1$--$L_2$ dictionary, then restructures translations using grammar rules which have been created by experts or are extracted from corpora. We propose a new method for phrase translation using an $L_1$--$L_2$ dictionary and n-gram data in $L_2$, instead of grammar rules, with a case study in translating phrases from Vietnamese to English. We note that the given Vietnamese phrases for translation do not exist in the dictionary. For example, we translate Vietnamese phrases ``{\fontencoding{T5}\selectfont b\d\ocircumflex} {\fontencoding{T5}\selectfont m\ocircumflex n khoa h\d{o}c m\'ay t\'inh}'', ``{\fontencoding{T5}\selectfont thu\'\ecircumflex} {\fontencoding{T5}\selectfont thu nh\d\acircumflex p c\'a nh\acircumflex n}'' and ``{\fontencoding{T5}\selectfont \dj\d\ohorn i m\d\ocircumflex t ch\'ut}'' to English: ``computer science department'',  ``individual income tax'', and ``wait a little'', respectively. In particular, given a Vietnamese phrase, our algorithms return a list of ranked translations in English.

One purpose of the phrase translations in our work is to support language learners. Assume that, using a Vietnamese-English dictionary, a learner has looked up translations of  ``{\fontencoding{T5}\selectfont b\d\ocircumflex} {\fontencoding{T5}\selectfont m\ocircumflex n'', ``khoa h\d{o}c'' and ``m\'ay t\'inh}'' as ``department/faculty'', ``science'' and ``calculator/computer'', respectively. Now, he wants to obtain the translation of ``{\fontencoding{T5}\selectfont b\d\ocircumflex} {\fontencoding{T5}\selectfont m\ocircumflex n khoa h\d{o}c m\'ay t\'inh}'', a phrase which does not exist in the dictionary. We present a method to generate phrase translations based on information in the dictionary.

\section{Overall Vietnamese morphology}
Vietnamese is an Austroasiatic language \cite{Lewis2014} and does not have morphology \cite{Thompson1963} and \cite{Aronoff2011}. In Vietnamese, whitespaces are not used to separate words. The smallest meaningful part of Vietnamese orthography is a syllable \cite{Ngo2001}. Some examples of Vietnamese words are shown as following:

\begin{itemize}
\itemsep0em
\item[--] Single words:  ``nh\`a''- house,  ``{\fontencoding{T5}\selectfont l\d{u}a}''- silk,  ``{\fontencoding{T5}\selectfont  nh\d\abreve t}''- pick up, ``mua''- buy and ``b\'an''- sell. 

\item[--] Compound words: ``mua b\'an''- buy and sell, ``b\`an {\fontencoding{T5}\selectfont gh\'\ecircumflex}''- table and chair, ``{\fontencoding{T5}\selectfont \dj\`\ocircumflex ng ru\d\ocircumflex
 ng}''- rice field, ``m\`e {\fontencoding{T5}\selectfont \dj}en''- black sesame,  ``{\fontencoding{T5}\selectfont  c\acircumflex y c\'\ocircumflex i}''- trees,  ``{\fontencoding{T5}\selectfont  \dj\uhorn\`\ohorn ng x\'a}''- street, ``{\fontencoding{T5}\selectfont m\~\acircumflex u gi\'ao}''- kindergarden, ``h\`anh ch\'anh''- administration, ``{\fontencoding{T5}\selectfont  th\h\ocircumflex}  {\fontencoding{T5}\selectfont c\h\acircumflex m}''- brocade, ``v\`ang v\`ang''- yellowish, ``{\fontencoding{T5}\selectfont ng\d{a}i ng\d{a}i}''- hesitate, ``{\fontencoding{T5}\selectfont g\d\acircumflex t g\`a g\d\acircumflex t g\`u}''- nod repeatedly out of satisfaction, ``{\fontencoding{T5}\selectfont l\h{a}i nh\h{a}i''- annoyingly insistent}.

\end{itemize}
Thus, what we call a \emph{word} in Vietnamese may consist of several syllables separted by whitespaces.

 \section{Related work}

The two methods, commonly used for phrase translation, are dictionary-based and corpus-based. A dictionary-based approach, e.g., \cite{Abiola2014} generate translation candidates by translating the given phrase to the target language using a bilingual dictionary. The candidates are restructured using grammar rules which are developed manually or learned from a corpus. In corpus-based approaches, a statistical method is used to identify bilingual phrases from a comparable or parallel corpus \cite{Pecina2008}, \cite{Koehn2003a}, and \cite{Bouamor2012}. Researchers may also extract phrases from a given monolingual corpus in the source language and translate them to the target language using a bilingual dictionary \cite{Cao2002}, and \cite{Tanaka2003}. Finally, a variety of methods are used to rank translation candidates. These include counting the frequency of candidates in a monolingual corpus in the target language, standard statistical calculations \cite{Pecina2008}, or even using Na\"ive Bayes Classifiers and TF-IDF vectors with the EM algorithm \cite{Cao2002}. \cite{Marino2006} extract translations from a bilingual corpus using an n-gram model augmented by additional information,  target-language model, a word-bonus model and two lexicon models.

More pertinent to our work is \cite{Hai1997}, who introduced a phrase transfer model for Vietnamese-English machine translation focusing on one-to-zero mapping, which means that a word in Vietnamese may not have appropriate single-word translation(s) and may need to be translated into a phrase in English. They translate Vietnamese words to English using a bilingual dictionary, then use conversion rules to modify the structures of the English translation candidates. The modifying process builds phrases level-by-level from simple to complex, restructures phrases using a syntactic parser and additional rules, and applies measures to solve phrase conflict.

\section{Proposed approach}

This section introduces a new simple and effective approach to translate from Vietnamese to English using a bilingual dictionary and n-gram data. An entry in n-gram data is a 2-tuple $<w_E, frq>$, where $w_E$ is a sequence of $n$ words  in English and $frq$ is the frequency of $w_E$. An entry in a bilingual dictionary is also a 2-tuple $<w_s,w_t>$, where $w_s$ and $w_t$ are a word or a phrase in the source language and its translation in the target language, respectively. If the word $w_s$ has many translations in the target language, there are several entries such as $<w_s,w_{t1}>$, $<w_s,w_{t2}>$ and $<w_s,w_{t3}>$. We note that an existing bilingual dictionary may contain phrases and their translations. Our work finds translations for phrases which do not exist in the dictionary. The general idea of our approach is that we translate each word in the given phrase to English using a Vietnamese-English dictionary, then use n-gram data to restructure translations. Our work is divided into 4 steps: segmenting Vietnamese words, filtering segmentations, generating ad hoc translations, selecting the best ad hoc translation, and finding and ranking English translation candidates.   

\subsection{Segmenting Vietnamese words}
A Vietnamese phrase \emph{P}, consisting of a sequence of $n$ syllables $<s_1$ $s_2$ ... $s_n>$, can be segmented in different ways, each of which will produce a segmentation \emph{S}. A segmentation \emph{S} is defined as an ordered sequence of words $w_i$ separated by semicolons ``;'' such as $S= <w_1; w_2; w_3;...; w_i;...;w _m>$, where $m$ is the number of words in $S$, $m \leq n$ and $1\leq i\leq m$. We note that a word may contain one or more syllables $s$. Generally, we have $2^{n-1}$ possible segmentations for a Vietnamese phrase \emph{P}. For example, the phrase ``{\fontencoding{T5}\selectfont khoa khoa  h\d{o}c}'' - science department/faculty, has 4 possible segmentations:\\
$S_1$ = <{\fontencoding{T5}\selectfont khoa; khoa; h\d{o}c}>, $S_2$ = <{\fontencoding{T5}\selectfont khoa; khoa  h\d{o}c}>,\\
$S_3$ = <{\fontencoding{T5}\selectfont khoa khoa;  h\d{o}c}>, and $S_4$ = <{\fontencoding{T5}\selectfont khoa khoa  h\d{o}c}>.

\subsection{Filtering segmentations}
Each word in each segment may have $k \geq 0$  translations in English. The total number of English translation candidates for a Vietnamese phrase, with $m$ words, is $O(2^{n-1}* m^k)$. To reduce the number of candidates, we check whether or not every Vietnamese word in each segmentation has an English translation in a Vietnamese-English dictionary. If at least one word does not have a translation in the dictionary, we delete that segmentation. For example, we delete $S_3$ and $S_4$ because they contain the words ``khoa khoa'' and ``{\fontencoding{T5}\selectfont khoa khoa  h\d{o}c}'' which do not have translations in the dictionary. As a result, the phrase ``{\fontencoding{T5}\selectfont khoa khoa  h\d{o}c}'' has 2 remaining segmentations: $S_1$=<{\fontencoding{T5}\selectfont khoa; khoa; h\d{o}c}> and $S_2$=<{\fontencoding{T5}\selectfont khoa; khoa  h\d{o}c}>. 

\subsection{Generating ad hoc translations}
\label{creatTrans}
To generate an ad hoc translation \emph{T}, we translate each word in a segmentation \emph{S} to English using the Vietnamese-English dictionary. The ad hoc translations of a given phrase are the translations of segmentations. For instance, the translations of the segmentation $S_1$ for ``{\fontencoding{T5}\selectfont khoa khoa h\d{o}c}'' are <faculty; faculty; study>, <department; department; study>, <subject of study; subject of study; study>; and the translations for $S_2$ are <faculty; science>, <department; science>, <subject of study; science>. Therefore, the six ad hoc translations of  ``{\fontencoding{T5}\selectfont khoa khoa h\d{o}c}'' are $T_1$=``faculty faculty study'', $T_2$=``department department study'', $T_3$=``subject of study subject of study study'', $T_4$=``faculty science'', $T_5$=``department science'', and $T_6$= ``subject of study science''.

\subsection{Selecting the best ad hoc translation}
We have generated several ad hoc translations by simply translating each word in the segmentations to English. Most are not grammatically correct. We use a method, presented in Algorithm \ref{alg:selectCandidate}, to reduce the number of ad hoc translations. We consider words in each entry in the English n-gram data as a bag of words $NB$ (lines 1-3), i.e., the words in each entry is simply considered a set of words instead of a sequence. For example, the 3-gram ``computer science department'' is considered as the set \{computer, science, department\}. Each ad hoc translation $T$, created in Section \ref{creatTrans}, is also considered a bag of words $TB$ (lines 4-6). For every bag of words $TB$, we find each bag of words $NB'$, belonging to the set of all $NB$s, such that $NB'$ contains all words in $TB$ (lines 7-9), i.e., $ TB \subseteq NB'$. Each bag of words $TB$ is given a score $score_{TB}$ which is the sum of frequency of all bags of words $NB'$ (line 10). The bag of words $TB$ with the greatest score is considered the best ad hoc translation (lines 12-18). 
\begin{algorithm}[!h]
\caption{Selecting the best ad hoc translation}
Input: all ad hoc translations $T$s\\
Output: the best ad hoc translation $bestAdhocTran$

\begin{algorithmic}[1]
\FORALL {entries $N \in$ n-gram data}
\STATE generate bags of words $NB$
\ENDFOR

\FORALL {ad hoc translations $T$}
\STATE generate bags of words $TB$
\ENDFOR

\FORALL {$TB$}
\STATE $score_{TB}=0$
\STATE Find all $NB'\in$ set of all $NB$s that contain all words in $TB$
\STATE $score_{TB}=\sum{Frequency(NB')}$
\ENDFOR

\STATE $bestAdhocTran$=$TB_0$
\FORALL{$TB$}
\IF{$score_{TB}>score_{bestAdhocTran}$}
\STATE $bestAdhocTran$=$TB$
\ENDIF
\ENDFOR
\STATE return $bestAdhocTran$

\end{algorithmic}
 \label{alg:selectCandidate}
\end{algorithm}

After this step, only one ad hoc translation $T$ will remain. For example, we eliminate 5 ad hoc translations (viz., $T_1$, $T_2$, $T_3$, $T_4$ and $T_6$) of the Vietnamese phrase ``{\fontencoding{T5}\selectfont khoa khoa h\d{o}c}'', and select ``department science'' ($T_5$) as the best ad hoc translation of it. We note that the best ad hoc translation may still be grammatically incorrect in English.
 
\subsection{Finding and ranking translation candidates}
To restructure translations, we use n-gram data instead of grammar rules. We take advantage that the n-gram information implicitly ``encodes'' the grammar of a language. Having the best ad hoc translation $TB$ and several corresponding bags $NB'$ from the previous step, we find and rank the translation candidates. For every $NB'$, we retrace its corresponding entry in the n-gram data, and mark the words in the entry as a translation candidate $cand$. Then, we rank the selected translation candidates.
\vspace{-2mm}
\begin{itemize}
\item If there exists one or many $cand$s such that the sizes of each $cand$ and $TB$ are equal, these $cand$s are more likely to be correct translations than other candidates. We simply rank $cand$s based on their n-gram frequencies. The candidate $cand$ with the greatest frequency is considered   the best translation. For example, the best ad hoc translation of ``{\fontencoding{T5}\selectfont khoa khoa h\d{o}c}'' is ``department science''. In the n-gram data, we find an entry <``science department'', 112> which contains exactly the same words in the best ad hoc translation found. We accept ``science department'' as a correct translation of ``{\fontencoding{T5}\selectfont khoa khoa h\d{o}c}'' and its rank is 112, which is the n-gram frequency of ``science department'.

\item The rest of the candidates are ranked using the following formula:
\vspace{-4mm}
\begin{center}
$rank(cand)= \frac{Frequency(cand)}{|size(cand)-size(TB)|*100}$.
\end{center} 

Our motivation for the rank formula is the following. If a candidate has a greater frequency, it has a greater likelihood to be a correct translation. However, if the size of the candidate  and the size of $TB$ are very different, that candidate may be inappropriate. Hence, we divide the frequency of $cand$ by the difference in the number of words between $cand$ and $TB$. To normalize, we divide results by 100.
\end{itemize}

\section{Experiments}
We work with the Vietnamese-English dictionary obtained from EVbcorpus\footnote{https://code.google.com/p/evbcorpus/}. The dictionary contains about 130,000 entries. We also use the free lists of English n-gram data available at the  ngrams.info\footnote{http://www.ngrams.info/} Website. The free lists have the one million  most frequent entries for each of 2, 3, 4 and 5-grams. The n-gram data  has been obtained from the corpus of contemporary American English\footnote{http://corpus.byu.edu/coca/}. 

Currently, we limit our experiments to translation candidates with equal or smaller than 5 syllables. We obtain 200 common Vietnamese phrases, which do not exist in the dictionary, from 4 volunteers who are fluent in both Vietnamese and English. Later, these volunteers are asked to evaluate our translations using a 5-point scale, 5: excellent, 4: good, 3: average, 2: fair, and 1: bad. 

The average score of translations created using the baseline approach, which is simply translating words in segments to English, is 2.20/5.00. The average score of translations created using our proposed approach is 4.29/5.00, which is quite high. The rating reliability is 0.72 obtained by calculating the Intraclass Correlation Coefficient \cite{Koch1982}.  
Our approach returns translations for 101 phrases out of the 200 input phrases. This means the precision and recall of our translations are 85.8\% and 50.5\%, respectively. 

We also compute the matching percentage between our translations and translations performed by the Google Translator. The matching percentage of our translations for phrases is 42\%. The translations marked as ``unmatched'' do not mean our translations are incorrect. A few such examples are presented in Table \ref{tab:CompareGoogle}.

\begin{table}[!h]
\centering
  \begin{tabular}{c}
\includegraphics[width=0.5\textwidth]{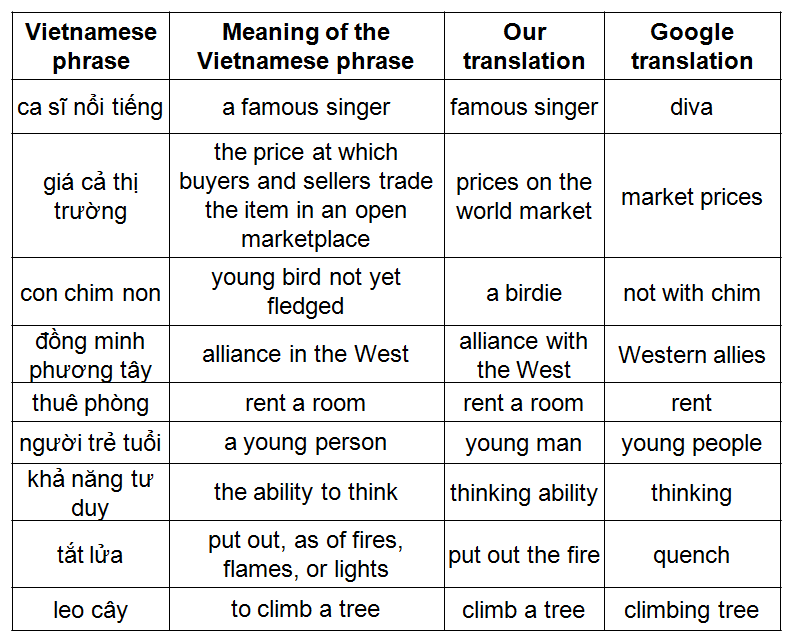}  
  \end{tabular}
\caption{Some translations we create are correct but do not match with translations from the Google Translator.}
\label{tab:CompareGoogle}
\end{table}  

The average score of our translations is high; however, the recall is low. If our algorithms can return a translation for an input phrase, that translation is usually specific, and is evaluated as excellent or good in most cases. Our approach relies on an existing bilingual dictionary and n-gram data in English. If we have a dictionary covering the most common words in Vietnamese, and the n-gram data in English is extensive with different lengths, we believe that our approach will produce even better translations.

\section{Conclusion and future work}
We have introduced a new method to translate a given phrase in Vietnamese to English using a bilingual dictionary and English n-gram data. Our approach can be applied to other language pairs that have a bilingual dictionary and n-gram data in one of the two languages. We plan to compute Vietnamese n-gram data from a Wikipedia dump and try to translate phrases from English to Vietnamese next.

\end{document}